\documentclass[10pt, conference, compsocconf]{IEEEtran}
\ifCLASSINFOpdf
\else
\fi
\usepackage[dvips]{graphicx}
\usepackage[latin1]{inputenc}
\usepackage{amssymb,amsmath,array}

\newtheorem{remark}{Remark}

\usepackage{color}
\begin{document}
%
\title{Short-term solar irradiance and irradiation forecasts \\ via different time series techniques: \\ A preliminary study}


\author{Cédric Join$^{1,2,3}$, Cyril Voyant$^{4,5}$, Michel Fliess$^{2,6}$, \\ Marc Muselli$^5$, Marie-Laure Nivet$^5$, Christophe Paoli$^7$ and Frédéric Chaxel$^1$
%
%
\vspace{.3cm}\\
%


1.- CRAN (CNRS, UMR 7039), Universit\'{e} de Lorraine \\ BP 239, 54506 Vand{\oe}uvre-l\`{e}s-Nancy, France
\\  {\tt \{cedric.join, frederic.chaxel\}@univ-lorraine.fr} \\
2.- AL.I.E.N. (ALgèbre pour Identification \& Estimation Numériques) \\ 24-30 rue Lionnois, BP 60120, 54003 Nancy, France \\
{\tt \{cedric.join, michel.fliess\}@alien-sas.com} \\
%
%
3.- Projet Non-A, INRIA Lille -- Nord-Europe, France \\
%
4.- Hôpital de Castelluccio, Unité de Radioth\'erapie \\ BP 85, 20177 Ajaccio, France \\
%
5.- SPE (CNRS, UMR 6134), Università di Corsica Pasquale Paoli \\ 20250 Corte, France \\
{\tt \{voyant, marc.muselli, marie-laure.nivet\}@univ-corse.fr} \\
6.- LIX (CNRS, UMR 7161), \'Ecole polytechnique \\ 91128 Palaiseau, France \\
{\tt Michel.Fliess@polytechnique.edu} \\
7.- D\'epartement de Génie Informatique, Université Galatasaray  \\ No:36 34357 Ortaköy, Istanbul, Turkey \\
{\tt cpaoli@gsu.edu.tr}
}

\maketitle

\begin{abstract}
This communication is devoted to solar irradiance and irradiation short-term forecasts, which are useful for electricity production. Several different time series approaches are employed. 
Our results and the corresponding numerical simulations show that techniques which do not need a large amount of historical data behave better than those which need them, especially when those data are quite noisy. 

\end{abstract}


%
\IEEEpeerreviewmaketitle

\section{Introduction}
\subsection{Generalities}\label{gener}
The aim of this communication is twofold:
\begin{enumerate}
\item It is devoted to solar irradiance and irradiation forecasts during rather short time intervals. As already noted in several publications (see, \textit{e.g.}, \cite{vi}, and \cite{notton1,notton2}) such predictions turn out to be very useful for electricity production by some systems like thermal and photovoltaic ones, where responses to solar variations are fast and complex. 
\item The above forecasts are achieved via different time series techniques which are compared. Although such an endeavor is of course not new (see, \textit{e.g.}, \cite{hossain,martin,rei}), this communication might be one of the very first which takes into account the implementation issues by evaluating quite closely their need of 
large historical data (see, also, \cite{hontoria}).
\end{enumerate}
Remember that times series (TS), which play an important rôle in numerical weather prediction  (see, \textit{e.g.}, \cite{duchon,hoca}, and the references therein), are also utilized in many other domains, like, for instance, econometrics and biology. We will not be employing here the classic setting for time series,\footnote{See \cite{meuriot} for a most interesting account and an analysis of the connections with econometrics and finance.} but
\begin{itemize}
\item  various topics from computer science, like artificial neural networks, computational intelligence and machine learning (see, \textit{e.g.}, \cite{beh,crone,de,kasa,kalo,mel,paoli,yona}).\footnote{The literature on the application to meteorological forecasts is already quite important. Combined with a lack of space it explains the absence of any review of those approaches in this communication.}
\item a new viewpoint on time series, which started in financial engineering (see, \textit{e.g.}, \cite{trend,agadir}), where
\begin{itemize}
\item no explicit mathematical model is needed,
\item the notions of \emph{trends} and \emph{quick fluctuations} are key ingredients.
\end{itemize}
\end{itemize}
Our time series were recorded every minute thanks to a meteorological station,\footnote{Vaisala Weather Transmitter WXT520 with an additive photosynthesis measuring probe head FLA613GS and the data loggers AHLBORN.}
which is on the roof of the \emph{Institut Universitaire de Technologie Nancy-Brabois}. In this paper, we are focusing on the solar irradiance in W/m$^2$ and irradiation in Wh/m$^2$. 


\subsection{Overview of the various techniques}
Our TS analysis may be divided  into two groups, \textit{i.e.}, with or without a need of big data.
\subsubsection{Settings without large historical data}
\begin{itemize}
\item The setting \emph{without model} (WM), which was sketched in Section \ref{gener}, should be related from an engineering viewpoint to the success of \emph{model-free control} \cite{csm}.\footnote{It might be interesting here to emphasize here that this viewpoint is also useful for greenhouses \cite{sousse}.}
\item The \emph{persistence} (P) method \cite{paoli}, which is a trivial machine learning viewpoint, assumes no change between the forecast and the last measure.
\end{itemize}
\subsubsection{Settings with large historical data}
A rather precise modeling is needed, which is quite greedy. We are considering four cases:
\begin{itemize}
\item A \emph{MultiLayer Perceptron} (MLP), which is a standard artificial neural network, is among the most popular tool for analyzing meteorological TS (see, \textit{e.g.}, \cite{beh,crone,v1}).
\item If we take into account a probabilistic description of the TS, a stationary hypothesis is often necessary. It yields a CSI-MLP, \textit{i.e.}, a \emph{clear sky index} in connection with the MLP in order to deal with stationary TS \cite{v2}. 
The broadband Solis model \cite{solis} is used: it allows generating the global irradiation without clouds. The ratio between measurement and clear sky solar radiation defines the clear sky index.
\item The \emph{scaled persistence} (SP) \cite{v2} adds to the classical persistence P a clear sky procedure like the above one. This methodology is equivalent to clear sky index persistence
\end{itemize}
After a short presentation of time series without model in Section \ref{AI}, various experiments are reported in Section \ref{exp}. Some concluding remarks may be found in Section \ref{conclusion}.

\section{Time series without any explicit model}\label{AI}
\subsection{Nonstandard analysis and the Cartier-Perrin theorem}
Take the time interval $[0, 1] \subset \mathbb{R}$ and introduce as
often in \emph{nonstandard analysis}\footnote{See, \textit{e.g.}, \cite{diener1,diener2} and \cite{lobry} for an introduction to this fascinating domain which is stemming from mathematical logic.} the infinitesimal sampling
$${\mathfrak{T}} = \{ 0 = t_0 < t_1 < \dots < t_N = 1 \}$$
where $t_{\iota + 1} - t_{\iota}$, $0 \leq \iota < N$, is {\em
infinitesimal}, {\it i.e.}, ``very small'' (\cite{diener1,diener2}).  A \emph{time series} $X(t)$ is a function $X:
{\mathfrak{T}} \rightarrow \mathbb{R}$.
\begin{remark}
The reader, who is not familiar with nonstandard analysis, should not be afraid by the wording \textit{infinitesimal sampling}. It just means in plain words that the sampling time interval is ``small'' with respect to the total  
recording time. Let us also stress that several time scales are most natural within this formalism.
 
\end{remark}

The {\em Lebesgue measure} on ${\mathfrak{T}}$ is the function
$\ell$ defined on ${{\mathfrak{T}}} \backslash \{1\}$ by
$\ell(t_{i}) = t_{i+1} - t_{i}$. The measure of any interval $[c, d]$, $0 \leq c \leq d \leq 1$, is its length $d -c$.  The
\emph{integral} over $[c, d]$ of the time series $X(t)$ is the sum
$$\int_{[c, d]} Xd\tau = \sum_{t \in [c, d]} X(t)\ell(t)$$
$X$ is said to be $S$-{\em integrable} if, and only if, for any
interval $[c, d]$ the integral $\int_{[c, d]} |X| d\tau$ is
\emph{limited}, \textit{i.e.}, not infinitely large, and also infinitesimal, if $d - c$
is infinitesimal.

$X$ is $S$-{\em continuous} at $t_\iota \in {\mathfrak{T}}$ if, and
only if, $f(t_\iota) \simeq f(\tau)$ when $t_\iota \simeq
\tau$.\footnote{$a \simeq b$ means that $a - b$ is infinitesimal.}
$X$ is said to be {\em almost continuous} if, and only if, it is
$S$-continuous on ${\mathfrak{T}} \setminus R$, where $R$ is a {\em
rare} subset.\footnote{See \cite{cartier} and \cite{lobry} for this technical definition.}
$X$ is \emph{Lebesgue integrable} if, and only if, it is
$S$-integrable and almost continuous.

A time series ${\mathcal{X}}: {\mathfrak{T}} \rightarrow \mathbb{R}$
is said to be {\em quickly fluctuating}, or {\em oscillating}, if,
and only if, it is $S$-integrable and $\int_A {\mathcal{X}} d\tau$
is infinitesimal for any {\em quadrable} subset.\footnote{A set is
\emph{quadrable} \cite{cartier} if its boundary is rare.}

Let $X: {\mathfrak{T}} \rightarrow \mathbb{R}$ be a $S$-integrable
time series. Then, according to the Cartier-Perrin theorem
\cite{cartier}, the additive decomposition
\begin{equation}\label{decomposition}
\boxed{X(t) = E(X)(t) + X_{\tiny{\rm fluctuat}}(t)}
\end{equation}
holds where
\begin{itemize}
\item the \emph{mean} $E(X)(t)$ is Lebesgue integrable,
\item $X_{\tiny{\rm fluctuat}}(t)$ is quickly fluctuating.
\end{itemize}
The decomposition \eqref{decomposition} is unique up to an additive
infinitesimal.
\begin{remark}
See \cite{lobry} for a less demanding presentation of the Cartier-Perrin theorem.
\end{remark}
\begin{remark}\label{noise}
The above quick fluctuations should be viewed like corrupting noises in engineering  \cite{bruit}. Determining the trend is therefore similar to noise attenuation. According to the mathematical definition of quick fluctuations, this may be achieved by integrating on a short time window or, more generally, by any low pass filter. 
\end{remark}

\subsection{On the numerical differentiation of a noisy signal}\label{ins} 
Let us start with the first degree polynomial time function $p_1 (t)
= a_0 + a_1 t$, $t \geq 0$, $a_0, a_1 \in \mathbb{R}$. Rewrite
thanks to classic operational calculus (see, \textit{e.g.},
\cite{yosida}) $p_1$ as $P_1 = \frac{a_0}{s} +
\frac{a_1}{s^2}$. Multiply both sides by $s^2$:
\begin{equation}\label{1}
s^2 P_1 = a_0 s + a_1
\end{equation}
Take the derivative of both sides with respect to $s$, which
corresponds in the time domain to the multiplication by $- t$:
\begin{equation}\label{2}
s^2 \frac{d P_1}{ds} + 2s P_1 = a_0
\end{equation}
The coefficients $a_0, a_1$ are obtained via the triangular system
of equations (\ref{1})-(\ref{2}). We get rid of the time
derivatives, i.e., of $s P_1$, $s^2 P_1$, and $s^2 \frac{d
P_1}{ds}$, by multiplying both sides of Equations
(\ref{1})-(\ref{2}) by $s^{ - n}$, $n \geq 2$. The corresponding
iterated time integrals are low pass filters which, according to Remark \ref{noise}, attenuate the
corrupting noises. A quite short time window is sufficient for
obtaining accurate values of $a_0$, $a_1$. Note that estimating $a_0$ yields the trend according to Remark \ref{noise}.

The extension to polynomial functions of higher degree is
straightforward. For derivatives estimates up to some finite order
of a given smooth function $f: [0, + \infty) \to \mathbb{R}$, take a
suitable truncated Taylor expansion around a given time instant
$t_0$, and apply the previous computations. Resetting  and utilizing
sliding time windows permit to estimate derivatives of various
orders at any sampled time instant.

\begin{remark}
See \cite{easy,NumDiff} for more details.
\end{remark}

\subsection{Application of the above calculations}
A ``good'' forecast $X_{\text{est}}(t+T)$, at time $t + T$, $T > 0$, of the time series $X(t)$ may be written
$$\boxed{X_{\text{est}}(t+T)=X_{\text{trend}}(t)+[\dot{X}_{\text{trend}}(t)]_eT}$$ 
where $[\dot{X}_{\text{trend}}(t)]_e$ denotes the derivative which is estimated as indicated in Remark \ref{noise} and Section \ref{ins}. In this paper,  
$T= 60$ minutes or $15$ minutes.

Let us stress that 
\begin{itemize}
\item $X_{\text{trend}}$ and $\dot X_{\text{trend}}$ are calculated with different time windows, \textit{i.e.}, respectively $10$ minutes and $75$ minutes.
\item What we predict is the trend and not the quick fluctuations (see also \cite{trend,agadir}).
\end{itemize}
\section{Experiments}\label{exp}

\subsection{Data}
\begin{figure}
\centering
\includegraphics[scale=.49]{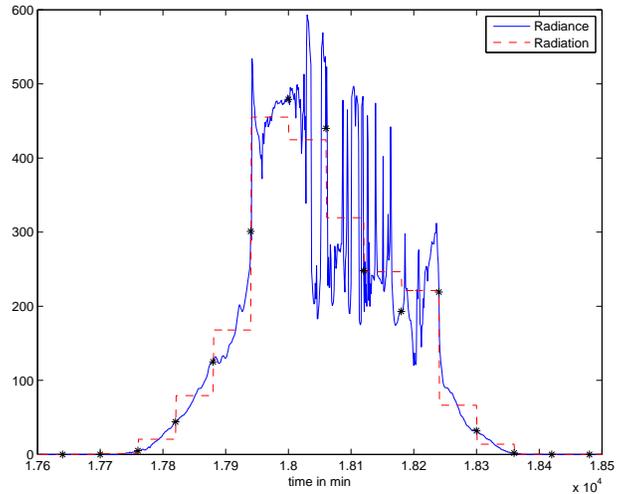}
\caption{Irradiance in W/m$^2$ (blue $-$), Irradiation in Wh/m$^2$ (red $- -$) and Irradiance each hour in W/m$^2$ (black $*$) on 13 February 2013}\label{res}
\end{figure}

We benefit from solar irradiance data, which were recorded during daylight every minute during three years, \textit{i.e.}, 2011, 2012 and 2013.
Two major types of experiments are distinguished:
\begin{enumerate}
\item irradiation is obtained according to the computation of the irradiance mean every hours, thus $\text{radiation}=\text{mean}(\text{radiance}(t-59), \text{radiance}(t-58),..., \text{radiance}(t))$ with $t=60 \times k$ minutes, $k=\{0, 1, 2, ...\}$.
\item Instantaneous measurements are used. 
\end{enumerate}
See the differences in Figure \ref{res}. Set $60$ minutes for the forecasting horizon.

\subsection{Irradiation forecasting}
\begin{figure}[h!]
\centering
\includegraphics[scale=.49]{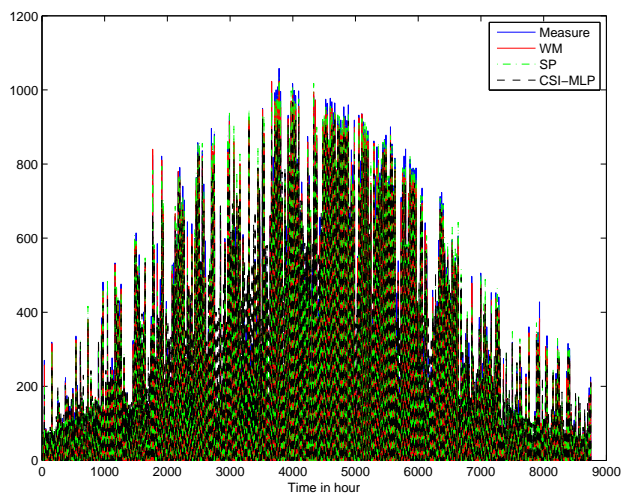}
\caption{Irradiation forecasting in Wh/m$^2$, year 2013}\label{ran}
\end{figure}

\begin{figure}[h!]
\centering
\includegraphics[scale=.49]{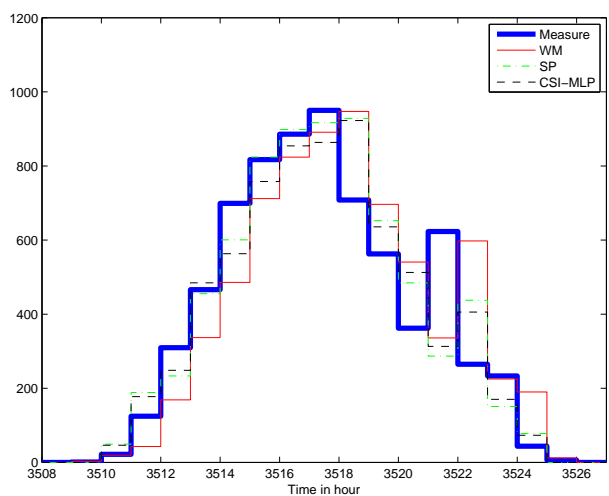}
\caption{Zoom of Figure \ref{ran}}\label{ranZ}
\end{figure}

All forecasting methods, namely P, SP, MLP and CSI-MLP, use directly irradiation measures, but not WM, for which records every minute permit better performances.
Figures \ref{ran} and \ref{ranZ} display the results for a $60$min forecast. More details may be found in the first lines of Tables \ref{TabL1} and \ref{TabL2}. The best method turns out to be SP. Even if the mathematical function ``clear sky'' plays a crucial rôle, his simplicity is remarkable. Note also that WM method is third with respect to the normalized $\text{L}^1$ norm. It becomes fourth with respect to the normalized $\text{L}^2$ norm.

\subsection{Irradiance forecasting}
\begin{figure}[h!]
\centering
\includegraphics[scale=.49]{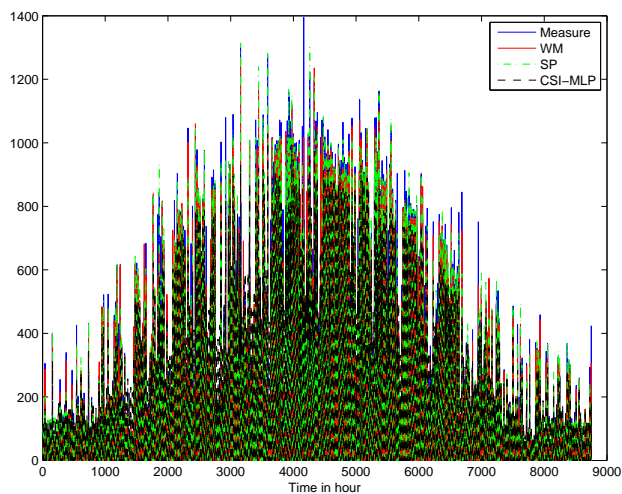}
\caption{Radiance forecasting in W/m$^2$, year 2013}\label{rad}
\end{figure}

\begin{figure}[h!]
\centering
\includegraphics[scale=.49]{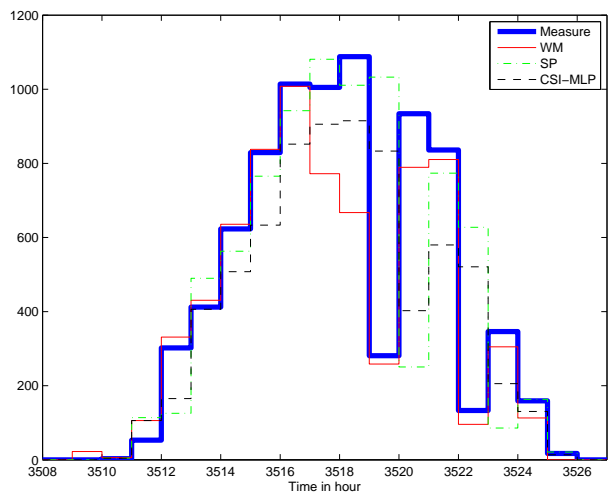}
\caption{Zoom on figure \ref{rad}}\label{radZ}
\end{figure}

We try now to forecast the irradiance, \textit{i.e.}, the instantaneous hourly values (see the stars $*$ in Figure \ref{res}). Contrarily to the previous calculations, averages are no more useful. Corrupting noises play therefore
a much more important rôle. 
Only instantaneous values are used for P, SP, MLP and CSI-MLP methods. For MLP and CSI-MLP, three learning data sizes were considered, \textit{i.e.}, one, two, and three years. The errors (normalized values) are related to the lower values computed among seven runs. Figures \ref{rad} and \ref{radZ} demonstrate that not only WM yields the best results but is also more robust with respect to the noises. WM is able moreover 
to take advantage of minute data: this a true advantage. Tables \ref{TabL1} and \ref{TabL2} confirm those statements.




\begin{table*}
  \centering
  \large
  \begin{tabular}{c|c|c||c|ccc|ccc}
\hline
 & P & WM & SP & \multicolumn{3}{c}{MLP} & \multicolumn{3}{c}{CSI-MLP}\\
 & & & & 1 year& 2 years& 3 years&1 year&2 years& 3 years\\
	\hline
Irradiation& 0.3373& 0.2393& 	{\bf 0.1962}& 0.2445 &0.2512 &0.2432 & 	0.2251&0.2240&0.2236\\
Irradiance& 0.4700& {\bf 0.2665}& 	0.3449& 	0.4705 & 0.4153 & 0.4177& 	0.3654&0.3544&0.3575\\
\hline
\end{tabular}
  \caption{Normalized $\text{L}^1$ norm of the five predictors for a $60$ minutes horizon and a time step of $60$ minutes. Best results in bold}\label{TabL1}
\end{table*} 

\begin{table*}
  \centering
    \large
  \begin{tabular}{c|c|c||c|ccc|ccc}
\hline
 & P & WM & SP & \multicolumn{3}{c}{MLP} & \multicolumn{3}{c}{CSI-MLP}\\
 & & & & 1 year& 2 years& 3 years&1 year&2 years& 3 years\\
	\hline
Irradiation& 0.6200& 0.4952& 	{\bf 0.44276}& 0.4763 &0.4751 &0.4690& 	0.4654&0.4626&0.44291\\
Irradiance& 1.0177& {\bf 0.5126}& 	0.8943& 	0.8860 & 0.8487 & 0.8447& 	0.7917&0.7872&0.7842\\
\hline
\end{tabular}
  \caption{Normalized $\text{L}^2$ norm for the five predictors for a $60$ minutes horizon and a time step of $60$ minutes. Best results in bold}\label{TabL2}
\end{table*}


\subsection{Extension to shorter time forecasts}
\begin{figure}[h!]
\centering
\includegraphics[scale=.49]{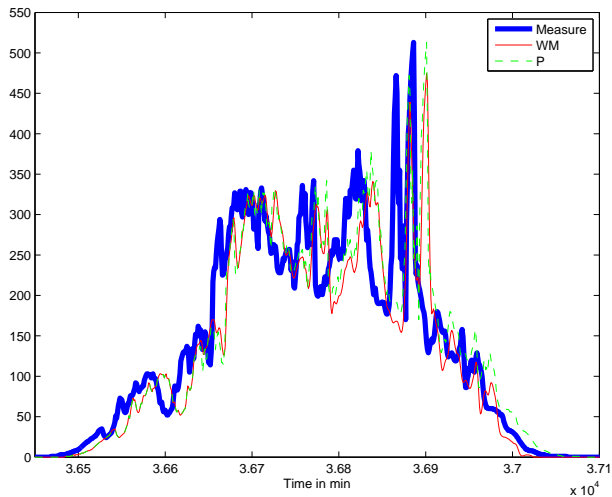}
\caption{Winter day (W/m$^2$)}\label{w}
\end{figure}

\begin{figure}[h!]
\centering
\includegraphics[scale=.49]{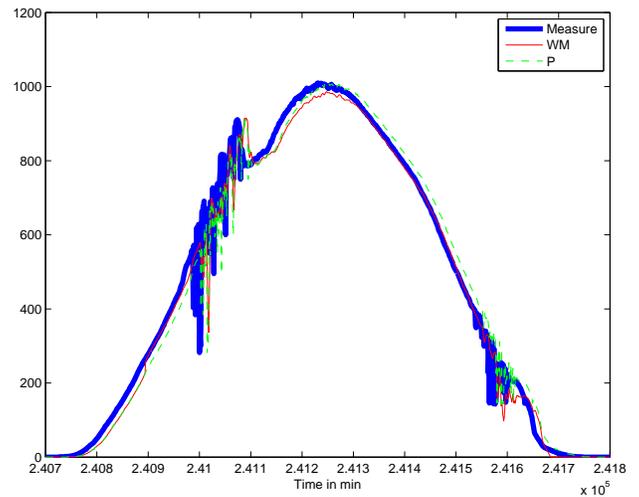}
\caption{Summer day (W/m$^2$)}\label{s}
\end{figure}

Switch now to a $15$ minutes forecast for the irradiation. The two most efficient methods are compared, \textit{i.e.}, SP and WM.
The normalized norm $\text{L}^2$ gives, for a time step of $5$ minutes and a $15$ minutes time horizon, respectively for SP and WM, $0.5624$ and $0.5399$. The superiority of the behavior of WM  may be given by the percentage $4.2 \%$.
Figures \ref{w} and \ref{s} show the predictor behavior predictor during one day in winter and in summer with slightly different time scales.

\section{Conclusion}\label{conclusion}
It is noteworthy to stress that methods with no need of a large amount of historical data give often better results, especially with noisy time series. Further investigations will hopefully confirm and precise those preliminary results. 
They should moreover yield new elements for epistemological discussions on the nature of ``good'' weather forecasts (see, \textit{e.g.}, \cite{murphy}).
\newpage
\section*{Acknowledgment}
The installation of the meteorological station was made possible by the project  E2D2, or \textit{\'Energie, Environnement \& D\'eveloppement Durable}, which is supported by the \emph{Universit\'e de Lorraine} and
the \textit{R\'egion Lorraine}.

\end{document}